# Depth analysis of battery performance based on a data-driven approach

*Zhen Zhang, Hongrui Sun, Hui Sun\**

State Key Laboratory of Heavy Oil Processing, College of New Energy and Materials, China University of Petroleum-Beijing, Fuxue Road No. 18, Changping District, Beijing 102249, PR China

**ABSTRACT:** Capacity attenuation is one of the most intractable issues in the current of application of the cells. The disintegration mechanism is well known to be very complex across the system. It is a great challenge to fully comprehend this process and predict the process accurately. Thus, the machine learning (ML) technology is employed to predict the specific capacity change of the cell throughout the cycle and grasp this intricate procedure. Different from the previous work, according to the WOA-ELM model proposed in this work ($R^2$ = 0.9999871), the key factors affecting the specific capacity of the battery are determined, and the defects in the machine learning black box are overcome by the interpretable model. Their connection with the structural damage of electrode materials and battery failure during battery cycling is comprehensively explained, revealing their essentiality to battery performance, which is conducive to superior research on contemporary batteries and modification.

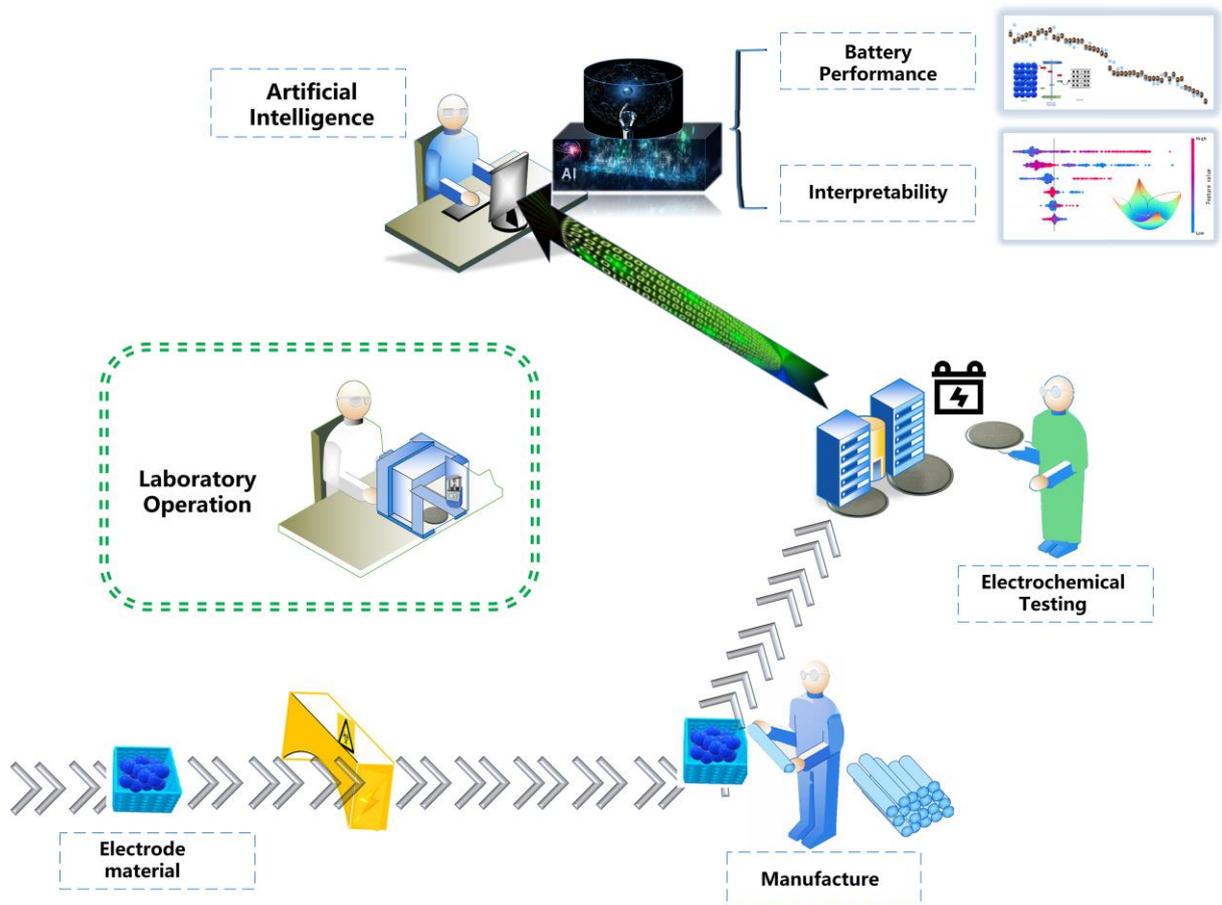

## 1. Introduction

With the aim of moderating the consumption of traditional fuels and carbon emissions, the vigorous development of the clean energy industry is currently a primary objective [1-3]. From the initial iron-nickel, lead-acid, alkaline batteries, and ultimately the most used lithium-ion batteries, it is evident that with the development and advancement of science and technology, the battery renews as quickly as possible [4,5]. Lithium-ion battery is a kind of rechargeable battery, which mainly by inserting Li-ions between the anode and the cathode to complete the charging and discharging operation. It is widely used for its economic cost advantages, high energy density and lifespan [6,7]. In recent years, with the emergence of Artificial Intelligence(AI), research into various materials has gradually shifted from traditional experimental methods to the paradigm of data-driven research. More and more researchers are using a combination of statistics and machine learning to predict physical properties [8,9], screening of new materials [10,11]. Coalescence of AI and batteries is highly practical because of the abundance of data and the availability of the batteries themselves. Meanwhile, state of health (SOH) estimation has caught the researchers' attention [12,13]. SOH is currently the most vital parameter for evaluating the health and performance of lithium-ion batteries, which is principally associated with battery aging models, because they track the authentic performance in operation, indicate the current capacity of the Li-ion battery, and describe the aging period of the battery [14]. While, for Li-ion batteries, the biggest obstacle in the application is the aging of the battery due to their step and complex aging mechanism, which depends on the operating environment and how the cell functions [15,16]. The capacity and power of will inevitably deteriorate because of it, thus decreasing its performance and cycle life, which is even worse causing some security risks [17-19]. It continues to be difficult to accurately quantify the whole process. Many experimental studies have been conducted to reveal the battery decay process, but since only one battery's performance tests often take hours or months, and the performance of different types and ratios of batteries varies greatly, it is both uneconomic and time consuming to repeat the experiments. As a result, the use of early cycle data to accurately predict life will create new occasions for battery production, utilization and majorization [20-23]. There are three broad classes of methods for estimating SOH: (1) Using mathematics, construct empirical and probabilistic models to simulate the decay process. (2) Physical and electrochemical procedures such as equivalent circuit models (ECM), filters. (3) Data-driven methods, especially machine learning (ML) algorithms, which have proven to be powerful and effective in helping to design and understand the decay process. In addition, a few studies have reported on the application of ML in the cell sector [24,25]. For example, S. Panchal et al. collected battery cycle data at four different temperatures and conducted a full-scale study, using the Thevenin degradation model to simulate [26]. Yang et al derived a detailed expression for the current time constant based on the first-order ECM, developed a prediction model for the current time constant τ based on a logarithmic function and a reference correlation curve to identify battery capacity decay [27]. Richardson et al based their in situ capacity estimation (GP-ICE) approach on smoothing the voltage profile with a Savitsky-Golay filter, and then used the voltage values as input to a Gaussian Process Regression (GPR) algorithm that operates on the voltage-time portion of the data itself, which is tantamount gap [28]. Luo et al designed a low-complexity model-based structure for estimating battery health by analyzing the errors of two RC ECMs and adapting to partitioning biases [29]. Shu et al identified the characteristics of the partial charging voltage profiles and used the Solid Least Squares Support Vector Machine (SL-SVM) model to estimate the battery SOH [30]. Yang et al. extracted characteristics of voltage, capacity and temperature profiles and applied the Gradient Lifting Regression Tree (GBRT) model to predict battery life with an average percentage error of approximately 7% [31]. The above shows demonstrates the potential application of ML to predicting battery performance.

However, existing research on battery SOH is still in the superficial layer of ML

applications, focusing only on predictive tasks and simple analysis of the importance of features. Of course, these efforts primarily aim to overcome the limitations of traditional methods, such as the high cost of trial and error and the inadequacy of data points. Furthermore, there are gaps in research in terms of explaining in depth the complex process of battery aging and understanding the interaction between operating conditions, testing practices, and battery performance. The process of battery disintegration is gradual as the number of battery cycles increases. The discharge process is more complicated and arduous to control, the charge is relatively easier to control in contrast, which will characterize the aging process appropriately. Theoretically, the more serious the aging of the battery, the shorter the charge and discharge time, and the lower charge/discharge capacity will be. From the existing research, we perceive that the decay process of the battery is roughly positively correlated with the trend in charge time, but not totally synchronized, so we need to go a little further to identify the reasons for it.

In this work, we have proposed a novel data-driven model which integrates a natural optimization algorithm and is accompanied by a structured data processing algorithm. This model doesn't need to take into account the complex mechanisms within the battery, but only the relationship between multiple input variables and the specific capacity of the battery. What's more, the time variation data of each voltage range under constant current charging condition are also collected, and the capacity attenuation of the battery is deeply interpreted from a new perspective.

## 2. Methods

*2.1 Experiment data.*

This paper has chosen to use a lithium iron phosphate battery with a nominal capacity of 170 mAh and a ternary battery with 200 mAh, collected by the individual's team from laboratory and online environments, as a case study to predict the battery lifetime. The dataset contains information on different aspects of the aging process, such as start voltage, end voltage, and charging time. Prior to each charge, the battery was left at rest for 2 hours to keep the condition of the battery as consistent as possible, and after the charge was completed, it was left at rest for 1h before being discharged. Each cycle is charged and discharged using a constant current mode and this data is collected directly from our laboratory for consistent programming and quality control.

*2.2 Exploratory data analysis.*

Metadata analysis of the data was conducted to gain a profound cognition of the battery aging process and to explain the importance of inputs in the prediction model. More specifically, the Pearson's correlation coefficient (PCCs) [32] and Grey Relation Analysis (GRA) [33] were used to reflect the correlation between inputs and outputs, to provide distinct consequences for features affecting outputs.

The basic idea of PCCs is to use the covariance and the standard deviation of the data to measure the correlation between the data, as shown in equation (1); GRA is a method used for describing and quantitatively comparing the dynamics of change in system development, and the basic idea is to determine whether the reference data column correlates with several columns of comparison data by determining the degree of difference in their geometry, as shown in equation (2)(3). In addition, the Shapley Additive Ex Planations (SHAP) analytical method, first proposed in 2016 by Ribeiro et al [34] and first applied in the battery domain in this work, has been used to overcome the 'black box' problem in machine learning and to give the interpretable model. The interpretation of the SHAP value is presented as an additive characteristic imputation method that interprets the predicted model value as the sum of the imputed values of each input characteristic, with the basic principle shown in equation (4).

$$r = \frac{\sum_{i=1}^{n}(X_i-\bar{X})-(Y_i-\bar{Y})}{\sqrt{\sum_{i=1}^{n}(X_i-\bar{X})^2}\sqrt{\sum_{i=1}^{n}(Y_i-\bar{Y})^2}}$$

(1)

r is the correlation coefficient with a value of $-1 \leqslant r \leqslant +1$. Besides the closer |r-| gets to 1, the closer the linear relationship between the two variables; The closer gets to 0, the weaker the linear correlation between the two variables. Generally, it may be divided into three levels: |r|<0.4 for low linear correlation; $0.4 \leqslant |r| < 0.7$ is significantly correlated; $0.7 \leqslant |r| < 1$ is highly correlated linearly.

$$\xi_i(k) = \frac{min_i min_k |x_0(k)-x_i(k)|-\rho \cdot max_i max_k |x_0(k)-x_i(k)|}{|x_0(k)-x_i(k)|+\rho \cdot max_i max_k |x_0(k)-x_i(k)|}$$

(2)

$$r_i = \frac{1}{n}\sum_{k=1}^{n}\xi_i(k), k = 1,2,...n$$

(3)

$\xi_i(k)$ is the correlation coefficient, $r_i$ is the relational degree, $\rho$ is identification factor, $r_i$ takes values within (0,1), with larger values indicating a stronger association. With respect to the identifying factor, the best resolution is obtained when $\rho \leq 0.5463$, usually taken as $\rho = 0.5$.

$$g(z') = \Phi_0 + \sum_{j=1}^{M} \Phi_j z'_j$$

(4)

$g(z')$ is the expository model, $z' \in \{0,1\}^M$ implies whether the homologous feature can be observed (1 or 0), It should be unstructured data, where the characteristics of each instance of structured data are observable (including missing values), M is the amount of input features, $\Phi_j \in R$ is the attribution value for each feature, $\Phi_0$ is the constant that illustrates the model.

*Feature Engineering.*

In this section, we removed features that could not help us in the relation between the features and the target based on the modeling results, returning and redoing the selection based on the significance of the features and the analysis of the relevance of the model. This is conducive to the model to better generalize span-new data and produce a more interpretable model. Feature engineering is an iterative process that generally requires multiple attempts at success. These processes require a mixture of expertise in the field of batteries and statistical data quality.

It is important to note that battery datasets tend to be high dimensional, often leading to dimensional disasters while calculating. Therefore, we use the non-linear algorithm t-distributed stochastic neighbor embedding (t-SNE) for feature fusion. This method improves on the primigenial SNE, which uses a symmetric loss function(5). $P_{j|i}$ denotes the odds of a high-dimensional space sample and $x_i$ choosing $x_j$ as its immediate neighbors, and the two-parametric distance is significantly negatively correlated to $P_{j|i}$. Let $P_{j|i}=P_{i|j}$, $q_{j|i}=q_{i|j}$, which are defined as (6)(7), here $\sigma^2$ is the variance of the Gaussian distribution centered on $x_i$. Because the Gaussian distribution is more sensitive to outliers, the fitting results for those with outliers deviate far from the real value and exhibit large variance, whereas for the t-distribution, it has better robustness to outliers and capture the overall situation of the data, and uses the t-distribution instead of the Gaussian distribution in the low-dimensional space, alleviating the space congestion perplexity of SNE, thereupon then effectively avoiding the dimensional catastrophe and also more convenient to analyze the correlation between inputs and outputs. To better determine the dimensionality of the fused features, screening is performed using Kullback-Leibler divergence (KL), [35] which is used to describe the difference between two probability distributions, defined as (8)

$$C = \sum_i KL(P_i \| Q_i) = \sum_i \sum_j P_{j|i} \log \frac{P_{j|i}}{q_{j|i}}$$

(5)

$$P_{i|j} = \frac{\exp(-\frac{\|x_i - x_j\|^2}{2\sigma^2})}{\sum_{k \neq l} \exp(-\frac{\|x_k - x_l\|^2}{2\sigma^2})}$$

(6)

$$P_{i|j} = \frac{\exp(-\frac{\|y_i - y_j\|^2}{2\sigma^2})}{\sum_{k \neq l} \exp(-\frac{\|y_k - y_l\|^2}{2\sigma^2})}$$

(7)

$$KL[P(X) \| Q(X)] = \sum_{x \in X} P(x) \log \frac{P(x)}{Q(x)} = E_{x \sim P(x)} \left[ \log \frac{P(x)}{Q(x)} \right]$$

(8)

For the voltage range and its corresponding time, which are the key indicators of this work, determinating of the voltage range is the most crucial part; on the one hand, the starting and finishing points and positions of the corresponding charge and discharge curves are different for diverse batteries because of the homologous materials and under the specific operating environment; on the other hand, the same type of battery ages progressively as the cycle progresses, and the starting and finishing points and positions of the corresponding charging curves are different. Fortunately, the shape of the charging curve remains almost aptotic, and there is a general tendency to move to the left during the process, as can be clearly seen in Fig. 1a. Since there is no mechanism to take into account the determinant effect of charging time in a given voltage segment on battery aging, this part will be the object of a detailed analysis in this work. We fit the charging curve of the entire cycle as the primary function line for easy visualization of curve changes. As can be seen in (Fig. 1b,c), the slope and interception of the fitted line change continually as the frequency of the cycles increases, with the slope being almost positively correlated with the number of cycles and the intercept changing in a wave-like pattern. After adjusting the slope and intercept of the whole process, which leads to the voltage study ranges VS_1, VS_2, VS_3 in this paper, as illustrated in Fig. 1d.

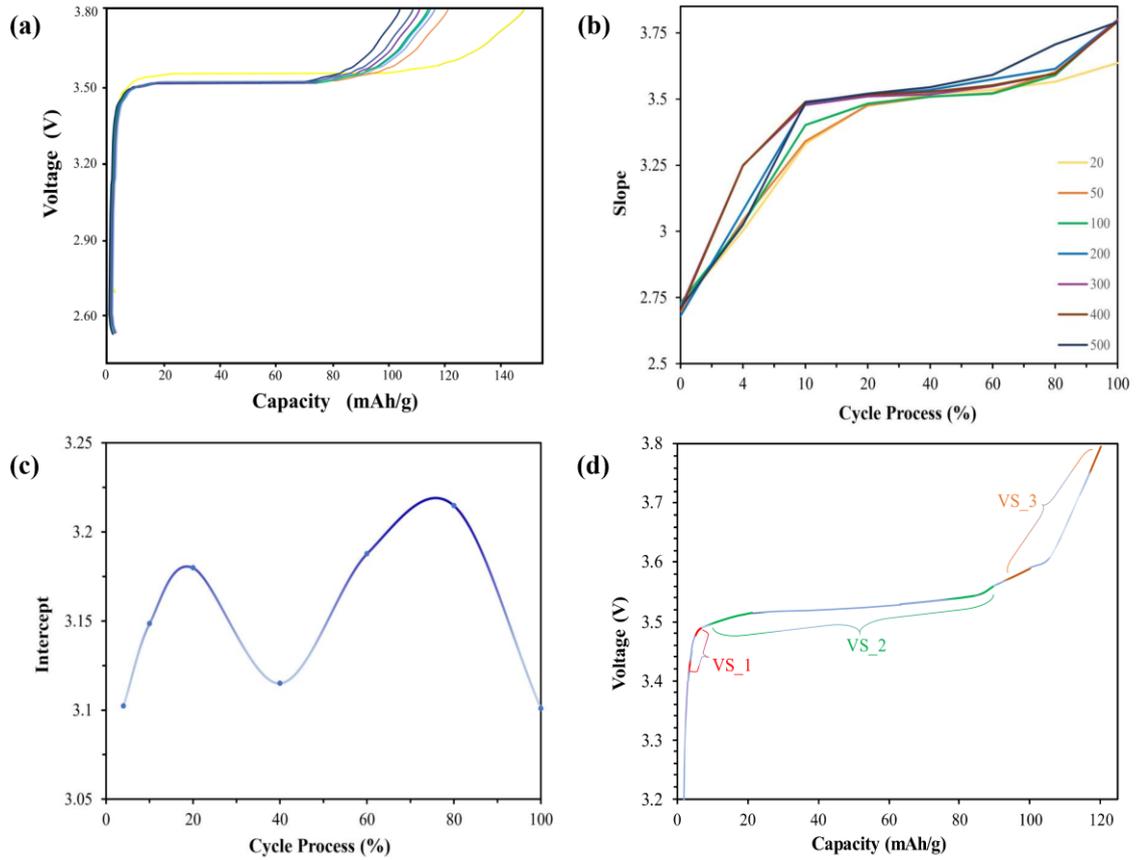

**Fig. 1.** (a) Charging curve trend of the battery throughout the cycle. (b) Slope after fitting to a linear over the whole cycle. (c) Intercept after fitting to a linear over the whole cycle. (d) voltage range obtained by combining the data distribution and curve inflection points. (a)(b) in which the colors from light to dark signify a gradual increase in the number of cycles.

*2.3 Model description.*

Extreme Learning Machines(ELM), proposed for the first time by Guangbin Huang in 2004 [36], is a special feed-forward neural network whose basic idea contains the randomizing connection weights of the input and hidden layers, the threshold of the hidden layer, without it being necessary to adjust the weights as a function of the back propagation of the gradient, cutting off a considerable part of the computational effort, while for the training process of forming ordinary neural networks, weights and deviations between neurons should be constantly adjusted for the data of the training set. The interface weight matrix ω between input and implicit layers is defined as:

$$\omega = \begin{bmatrix} \omega_{11} & \omega_{12} & \cdots & \beta_{1\,n-1} & \beta_{1n} \\ \omega_{21} & \omega_{22} & & \beta_{2\,n-1} & \beta_{2n} \\ \vdots & & \ddots & \vdots & \\ \omega_{l-1\,1} & \omega_{l-1\,2} & \cdots & \beta_{l-1\,n-1} & \beta_{l-1\,n} \\ \omega_{l1} & \omega_{l2} & & \beta_{l\,n-1} & \beta_{ln} \end{bmatrix}_{l \times n}$$

(9)

n is the number of neural input terminals in the input layer, b is the bias of the hidden layer, corresponding to l neural nodes in the hidden layer, and the bias vector $\vec{b}$ is a vector of 1×1 column:

$$\vec{b} = \begin{bmatrix} b_1 \\ b_2 \\ \vdots \\ b_l \end{bmatrix}_{l \times 1}$$

(10)

β is a matrix of l × m, representing the connection weights of the implied layer and the output layer; The initial implicit layer weights ω and bias b are used to solve the output layer weight matrix, allowing the ELM model to be obtained.

$$\beta = \begin{bmatrix} \beta_{11} & \beta_{12} & \cdots & \beta_{1\,m-1} & \beta_{1m} \\ \beta_{21} & \beta_{22} & & \beta_{2\,m-1} & \beta_{2m} \\ \vdots & & \ddots & \vdots & \\ \beta_{l-1\,1} & \beta_{l-1\,2} & \cdots & \beta_{l-1\,m-1} & \beta_{l-1\,m} \\ \beta_{l1} & \beta_{l2} & & \beta_{l\,m-1} & \beta_{lm} \end{bmatrix}_{l \times m}$$

(11)

The calculation is $f_L(x) = \sum_{i=1}^{L} \beta_i g(\omega_i \cdot x_i + b_i) = \sum_{i=1}^{L} \beta_i h(x_i)$

(12)

Here, g is an activation function, $h(x_i)$ is the mapping value of the i th sample feature, L is the total amount of samples, and H is the output matrix of the hidden layer. The above formula can be abbreviated as T=Hβ, let T be the objective matrix of the training set, so that only the appropriate β such that the error function value is minimum or close to 0.

$$\|H\beta - T\|^2 = 0$$

(13)

The Whale Optimization Algorithm (WOA). A variety of natural algorithms with intrinsic convergence and evolutionary characteristics have emerged over the last few years, mainly through optimizing strategies that simulate natural processes such as animal populations in nature, genetic algorithms, ant colony algorithms, particle swarm algorithms and grey wolf algorithms, which have an esoteric natural interpretation, both in terms of natural parallelism and a high degree of fault tolerance to the objective function. This paper uses WOA, an algorithm first proposed by Mirjaili in 2016, [38] which is a meta-heuristic majorization algorithm that simulates humpback whale hunting behavior using stochastic or optimal iterative research to simulate hunting behavior and utilizes the upward spiral model to simulate the bubble net feeding method of humpback whales: whales highlight spiral-shaped bubbles around their prey at the surface of the sea during feeding.

The WOA algorithm assumes that the currently adopted pattern is the closest to the target prey, and updates the information obtained after continuous iterations until the kill is hit. The predatory behavior is divided into three categories, the first of which is encircling the prey:

$$\vec{D} = |\vec{C} \cdot \vec{X}^*(t) - \vec{X}(t)|$$

(14)

$$\vec{X}(t+1) = \vec{X}^*(t) - \vec{A} \cdot \vec{D}$$

(15)

$$\vec{A} = 2\vec{a} \cdot \vec{r} - \vec{a}$$

(16)

$$\vec{C} = 2 \cdot \vec{r}$$

(17)

$$a = 2 - (\frac{2t}{t_{max}})$$

(18)

Here t is the current number of iterations, $t_{max}$ is the maximum number of iterations, $\vec{X}^*(t)$ is the current optimal search agent, $\vec{X}(t)$ is the position vector of the current solution, and D is the distance between the whale and the prey in the Nth dimension. During the iteration, if $\vec{X}(t+1)$ is better than $\vec{X}^*(t)$, then $\vec{X}^*(t)$ is automatically updated, with a random vector of $\vec{r}\epsilon[0,1]$, $\vec{A}$ and $\vec{C}$ are both coefficient vectors, and *a* is an iteration factor and decreases with increasing of iteration times. The second, the spiral model (upward-spirals), calculates the distance between the position of the whale (x,y) first to and the position of the prey (x*,y*) and then determines a spiraling equation between the two (19)(20).

$$\vec{X}(t+1) = \vec{D'} \cdot e^{bl} \cdot \cos(2\pi l) + \overrightarrow{X^*}(t)$$

(19)

$$\vec{D'} = |\overrightarrow{X^*}(t) - \vec{X}(t)|$$

(20)

$D'$ expresses the distance between the i th whale and its prey, b is a constant which restricts the form of the logarithmic spiral, and l is a random number between [-1,1].

The third, randomized selection, updates the position of the search agent in the scanning phase based on the selected randomness and allows the WOA algorithm to perform a global hunt.

$$\vec{D} = |\vec{C} \cdot \vec{X}_{rand} - \vec{X}| \qquad (21)$$

$$\vec{X}(t+1) = \vec{X}_{rand} - \vec{A} \cdot \vec{D} \qquad (22)$$

Here $\vec{X}_{rand}$ is a randomly picked whale.

## 2.4 WOA-ELM

In order to further improve the predictive precision of the specific capability of the battery, this work aims at optimising on the basis of the original ELM by integrating the whale optimization algorithm to obtain a brand new model (WOA-ELM) see Fig. 2. Because WOA is a metaheuristic algorithm, the new model can well overcome the slow learning speed and learning rate, and it is very facile to fall into the minimal local error. The model divides the dataset into a training set (70%) and a test set (30%). After iterating the model parameters across the training set, the final well-formed model is obtained by retraining using the qualified fitness value and the optimal individual solution.

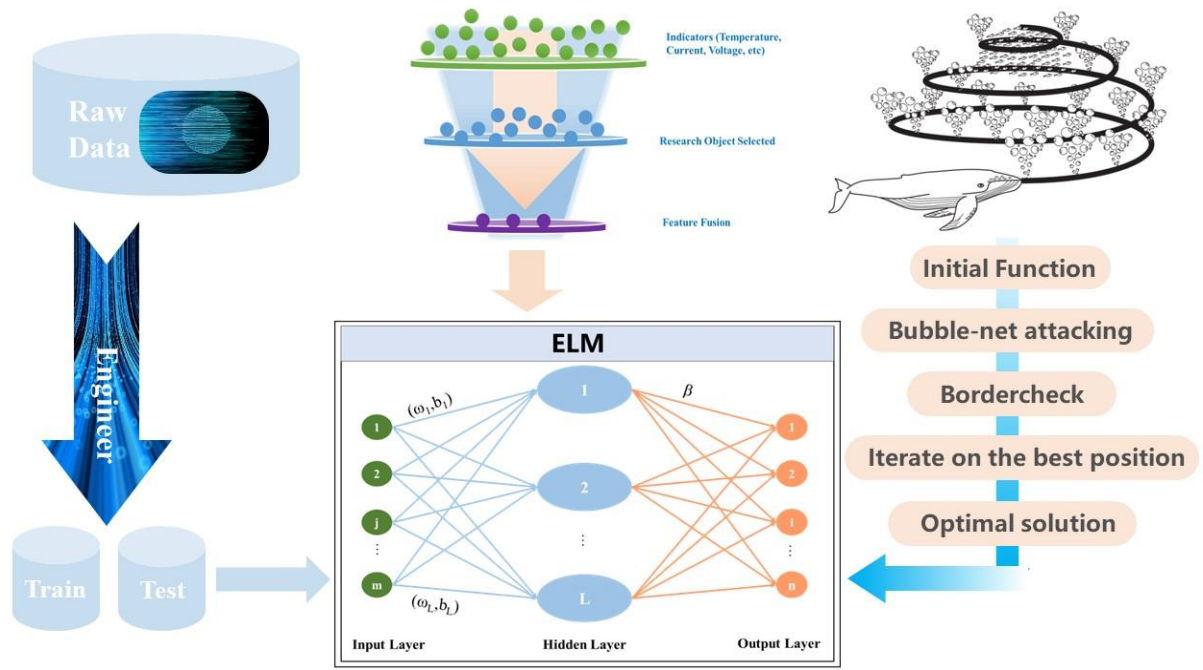

**Fig. 2.** WOA-ELM model flow chart.

*2.5 Model Evaluation*

To quantitatively evaluate the performance of ML models, Standard Deviation (SD), coefficient of determination ($R^2$), and root mean square error (RMSE) are used to represent the predictions accuracy and visualized using Taylor diagrams, where each point on the diagram represents the difference between the predicted and true values of the model, which represent the model's accuracy. Taylor diagrams allow a superior capability comparison between several models to evaluate the best model; SD reflects the degree of dispersal between individuals within the sample, and the larger the SD, the greater the difference between its real value and the average, defined as (23); RMSE is used to assess the local prediction accuracy, defined as equation (24); $R^2$ reflects the proportion of the variation of the dependent variable that the independent variable may account for by the regression relation, which can convey the adjustment notch between the estimated value of the trendline and the corresponding actual, the higher value indicates that the model is striking, the max is 1, defined as equation (25) here n is the number of samples.

$$\sigma_A = [\frac{1}{N}\sum_{n=1}^{N}(A_n - \bar{A})^2]^{\frac{1}{2}}$$

(23)

$$RMSE = \sqrt{\frac{\sum_{i=1}^{n}(Predict_i - Actual_i)^2}{n}}$$

(24)

$$R^2 = \sqrt{\frac{\sum_{i=1}^{n}(y_i - \bar{y})^2}{\sum_{i=1}^{n}(y_i - \bar{y})^2 + \sum_{i=1}^{n}(y_i - \hat{y})^2}}$$

(25)

**3 Results and discussion**

*3.1 Pivotal factors influencing battery performance.*

According to the WOA-ELM pattern manufactured in this work, the influence of features on the results is discussed qualitatively by exploring the GPA and PCC methods, and the cumulative rankings in the importance of the different input characteristics are similar for the two analytical methods for the two targets (Fig. 3 a,b). It is obvious that the three features significantly correlated with the output are F8, F12 and F13. The charging time corresponding to VS_2 shows a straightforward link with the specific capacity of the battery. The time interval of the VS_3 segment shows both a weak linear and monotonous relation to the specific capacity, and the trend of charge/discharge specific capacity is always the same.

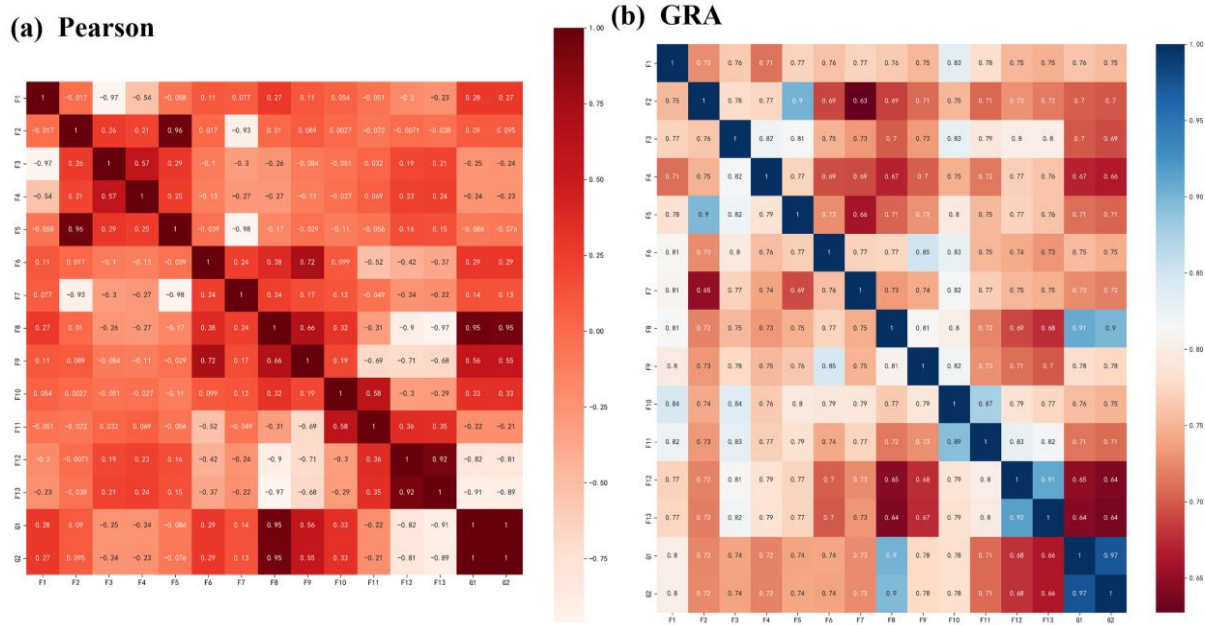

**Fig. 3.** (a) PCC results. (b) GRA results.

In combination with the (Random Forest) RF model, SHAP method was used to quantify the influence factors of the features by performing pre-order factor analysis on the dataset. SHAP interprets the predictive value of the model as the sum of the imputed values of each input, and if the SHAP value increases accordingly, the variable has a positive effect on the battery performance. As shown in Fig. 4a for the interpretation of individual predictions, the longer the unit of length, the greater the impact effect. Our baseline value is 97.35, red represents an optimistic influence on the target output, blue represents a passive impact on the target output, the impact of F8 is a negative correlation and affects the output from 97.9 to 97.26, the impact of F1 is a positive correlation increasing from 96.8 to 97.62. In Fig. 4b, the baseline value is 97.4 and the positive correlation of F8 for the output is marked and drives the output from 96.5 to 98.29. In the histogram of Fig. 4c we may obtain the rank of the degree of influence of the features on the output. In Fig. 4d, each point represents sample, the more visible the red, the greater the value of the element itself, and the same applies to red. From the collected dataset, it is apparently F8 which is the most influential characteristic, and the red points are concentrated on the right side, revealing that the larger the value of F8, the higher the specific capacity, which exerts a considerable positive influence on this one.

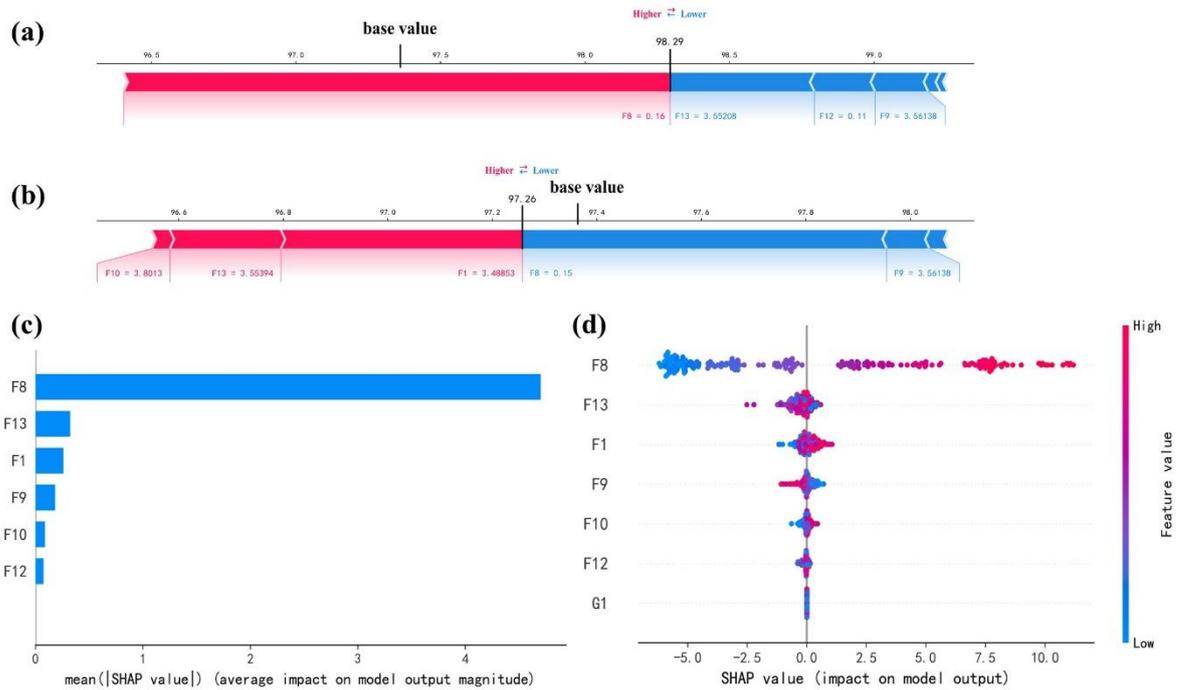

**Fig. 4.** (a) (b) Analysis of the projections for each sample. (c) Mean impact on the size model output. (d) SHAP values for individual features.

The result shows that comprehending the mechanism behind these factors can be of essential assistance in improving battery performance. For instance, the starting points of different segments of voltage have distinct effects on the specific capacity of the cell, both beneficial and detrimental, but too low starting voltages don't lead to a satisfactory level. Consequently, the correlation between the starting point of voltage stages and the specific capacity of the battery must be taken into account.

In addition to a significantly correlated target output, F8 also interacts with other input variables. The interactive analysis of multiple features is shown in Fig. 5. F8 and F13 have interaction, which corresponds to the knowledge of the battery itself. The LFP battery used in this work, the platform is more obvious, so the median voltage of the battery is the platform voltage. Moreover, we have also found that interaction between F1 and F12 is extremely weak, which is different from our previous knowledge, the level of the battery Initial voltage, is affected by the battery charging process. While it's too low will lead to charging process is incomplete, and Fig. 5 shows that we can independently consider the relationship between them.

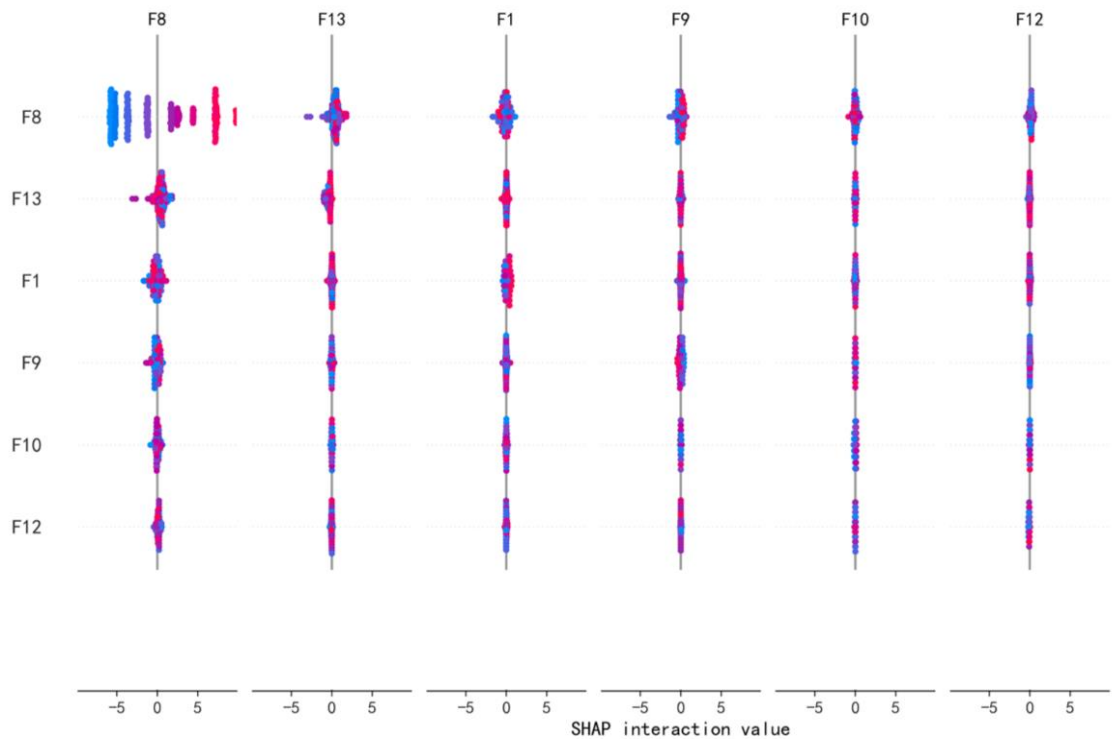

**Fig. 5.** Correlation of features and output.

*3.2 Battery forecast model.*

The results of this work are compared with the performance of various existing superior ML models SVM [38], RF [39], GBRT [40], K Near Neighbor(KNN) [41], Multilayer Perceptron (MLP) [42], ELM [36], As exhibited in Fig. 6a, the three evaluation indexes of SD, $R^2$, and RMSE of the model prediction results are constrained by the cosine function relationship using the Taylor diagram, and finally integrated into a polar graph, closer to the REF point, the eminent performance of the model is maintained, the dashed arc indicates the RMSE, and the r and θ axes represent the SD. It can be manifestly seen that the RF has an excessive RMSE, although the $R^2$ is high. While the WOA-ELM proposed in this paper has excellent performance in STD, $R^2$, and RMSE, with an $R^2$ of 0.999871, which is superior to the other six models. Fig. 6b shows the plot of predicted versus actual values as a function of the specific battery capacity at a 95% confidence level. RF and GBRT also have exceptional prediction performance with $R^2$ of 0.907551 and 0.844522 respectively. In (Fig. 6c, d), great accuracy can be reached for the majority of the points, and only a few points have errors. In spite of the training error tends to be inferior to the test error, which is a natural drawback of data-driven means, nevertheless the test performance of WOA-ELM prediction remains satisfactory and competitive with existing models.

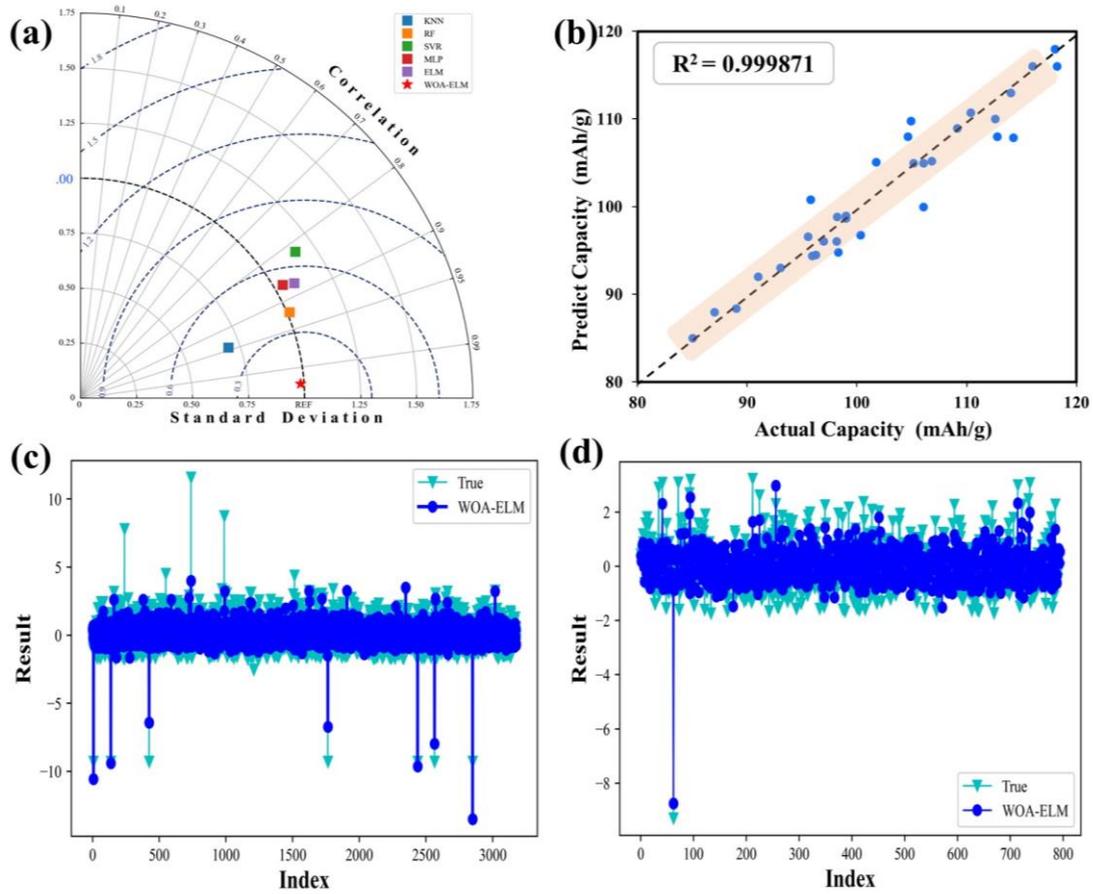

**Fig. 6.** (a) Talyor Diagram. (b) Predicted values contrast with actual values based on specific cell capacity at 95% confidence level. We plot the diagonal line (y = x) as a reference, and the closer the point of that diagonal line is, the higher prediction accuracy. (c)(d) Evaluation of the model training set and test set.

*3.3 The intrinsic connection of charging.*

In order to understand the whole process of battery charging, we start with have a concise understanding of the mechanism of the process as a whole. From the macroscopic point of view, in the trickle process, the internal reaction of the battery is slow, the polarization voltage is small, and the active material is distributed more evenly, and the battery is in the process of being gradually activated. All batteries will get this stage. Therefore, the time corresponding to the voltage range of the pre-charge stage has little influence on the final battery specific capacity, this work hasn't much research on the corresponding to the pre-charge voltage range too much. In the pre-charging stage, the power ought to slightly exceed the discharged amount, which has the greatest positive effect on the performance of the battery at present. Next is the standard charging phase: kinds of batteries have several chemical reactions with the various electrode and electrolyte materials used at this stage, so we explore the level of mechanism of the battery reaction, the polarization of the battery internal resistance increases abruptly, causing a rapid rise in voltage, and the battery voltage has reached the cut-off voltage prior to being fully charged. If you continue to choose to charge now, although you may continue to augment portion of the capacity, but to some certain extent will lead to the potential negative electrode battery is too low, resulting in the precipitation of lithium metal, the amount of active material in the battery is reduced and the internal resistance further increased. In according with above analyses, the charging time should be properly adjusted, otherwise if the time is too long, which will render lithium ions from the active material of the cathode with the fast speed movement to the anode and destruct the lattice framework.

  From a microscopic point of view, lithium-ions migrate from within the inside of the LFP crystal to the surface, enter the electrolyte under the action of electrical field force, then pass through the diaphragm, migrate to the surface of the graphite crystal via the electrolyte, embed in the graphite lattice lastly. Meanwhile, the electrons flow towards the cathode aluminium foil collector through the conductor body, and the flow towards the anode copper foil collector rest on the lug, pole, external circuit, then stream to the anode of graphite through the conductive body, thereby the electric charge of the anode reaches the balance anew. After the lithium ion is deintercalation from LFP, the LFP is converted into iron phosphate and a phase change occurs throughout the process. The proportion of the charging curve before the plateau VS_1: in the early stage when the $Li^+$ is less delithiated, the second phase is not nucleated, but converted from LFP to $Li_{1-\alpha}FePO_4$; charging plateau VS_2: the second phase is nucleated and forms a two-phase interface, and the material undergoes a two-phase reaction between the Li-rich phase $Li_{1-\alpha}FePO_4$ and the Li-deficient phase $Li_\beta FePO_4$ ($\alpha,\beta>0$); after the plateau VS_3: the Li-deficient phase has been formed, and the material produces $FePO_4$ by single-phase reaction [43].

From Fig. 4 we can see that the correlation between F8 and the specific capacity of the battery is much higher than the rest of the voltage range, and the above narrative explains that very clearcut conclusion. From Fig. 4d, F8 has the greatest influence on the specific capacity, which grows almost linearly with the amplification of the time corresponding to F8; while for the rest of the characteristics, there is some deviation always though, it may be approximately equal to the trend of specific capacity, which corresponds to the results of our mechanical analysis of the battery charging process in the preceding section.

*3.4 Solving dimensional disasters.*

The KL-divergence after feature fusion was calculated, because the smaller the KL, representing the smaller the difference in distribution between the data, the more similar the distribution is. The two-dimensional KL value is well above the one and three-dimensional value, as shown in Fig. 7a, which indicates a quantity of information in the original data can be retained in two dimensions. The dimension is thus determined as two-dimensional, which is most appropriate for further exploration. Because all of the existing dimensional reduction algorithms are of black box type. To overcome this problem and better understand how the original features contribute to the newly created features as shown in Fig. 7b, we investigate features that are pre and post fusion. Moreover, the speed and precision of calculation before and after the merging of functionalities were compared, and it was worth noting that the use of a small fraction of $R^2$ as a sacrifice has improved a large fraction of the computational speed, which is extremely significant for future processing and analysis of larger battery datasets, and the computational results in Table 1.

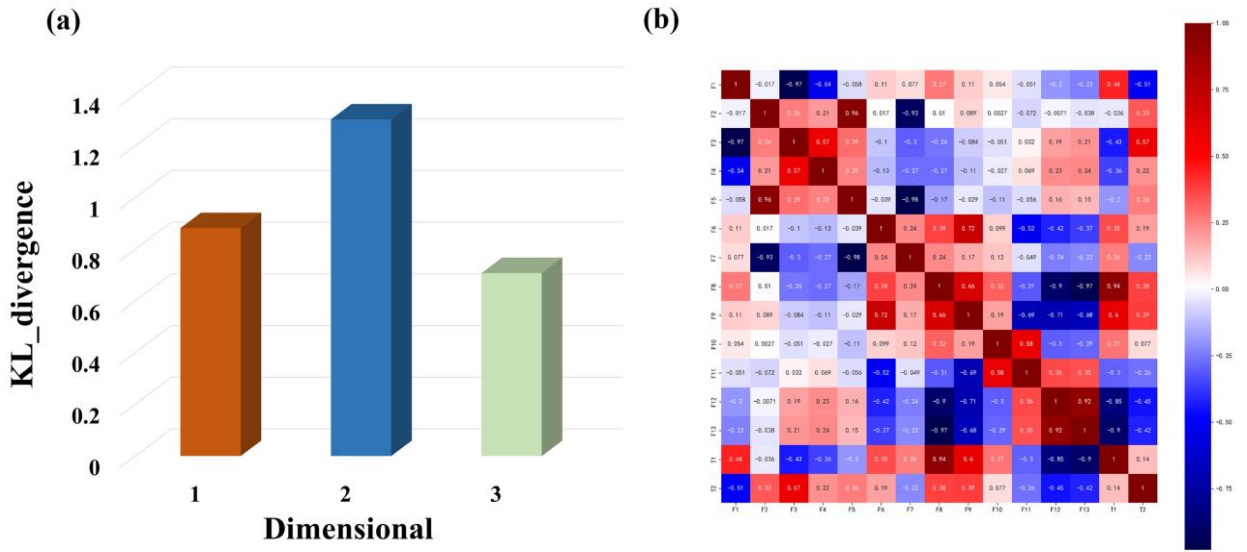

**Fig. 7.** (a) KL-divergence. (b) Correlation before and after feature fusion.

**Table 1**

Calculation results before and after feature fusion.

| Item | Before the fusion | After the fusion | Difference |
| --- | --- | --- | --- |
| RMSE | 0.1291 | 0.0015 | 98.84% |
| Training Data $R^2$ | 0.9998 | 0.9996 | -0.02% |
| Test Data $R^2$ | 0.9998 | 0.9994 | -0.04% |
| Time(mS) | 1556.6916 | 170.0368 | 89.08% |

*3.5 Software visualization*

In this study, the Python-PyQt5 module was used to produce the model as a graphical interface, making it more applicable. Fig. 8 shows the actual application interface of the software, which predicts the specific capability of the battery using an optimum WOA-ELM model preset when the user enters the appropriate eigenvalues. The actual battery data(that is not part of the training and testing data sets) in the laboratory has been tested to verify the high predictive accuracy of the model in the actual working state of the battery. The software is capable of generating accurate prediction results because it has a high predictive performance, using MSE to measure mean prediction error, data engineering characteristic for prediction, the software prediction time is less than 3 s. The software guides battery experiments and analyses the results of battery tests, whatever the condition in which the battery is tested.

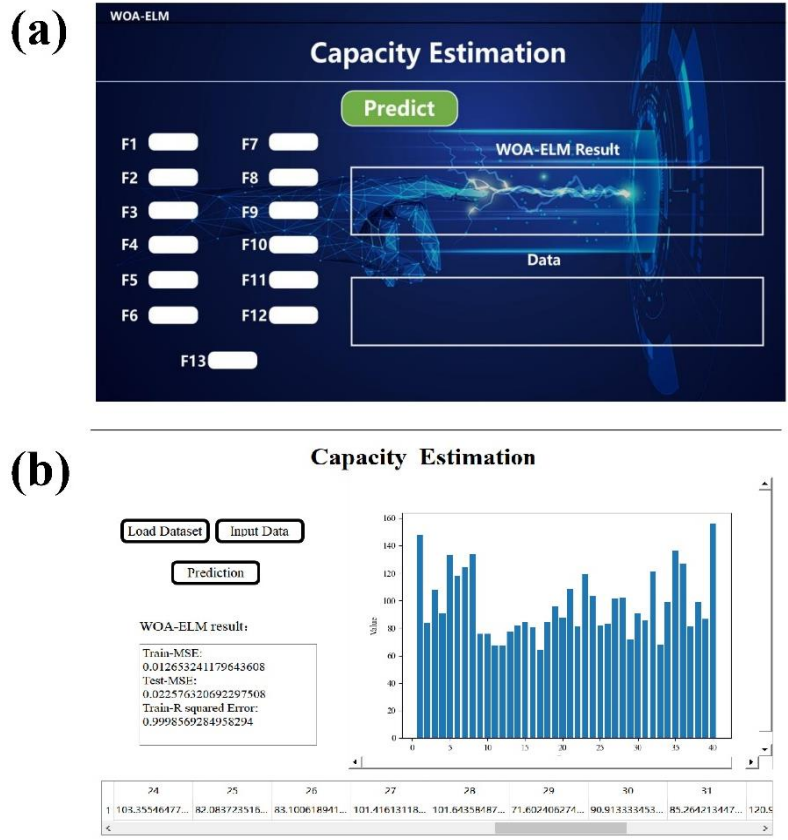

**Fig. 8.** (a) Software interface. (b) Display of running results.

## 4. Conclusion

This work provides a model framework based on machine learning combined with metadata analysis, which can effectively address the tricky problem of "black box" of traditional machine learning models. The whale optimization algorithm is used to construct the fitness function, determine the number of iterations, and select the reasonable population size, threshold and weights and other main parameters. On this basis, the charging voltage range is divided, the corresponding charging time is analyzed, and the influence of each parameter of the charging process on the performance of the battery is discussed. Ultimately, the effect of the corresponding time of different charging stages on the target output is interpreted in detail with explained from the level of mechanisation.

However, there are still some limitations in the present work, and the subtle effects of process conditions on the cell performance during the manufacturing process have not been discussed in this paper. Moreover, the optimal solutions obtained from the model calculations still require additional accuracy due to the non-negligible errors between the experiments and the calculations. After that, more data of different cell types with different process steps will be gathered about the manufacturing process (sintering time, sintering temperature, loading of active ingredients, modified materials) to expand the size of the dataset and meliorate the developed model. In the future, the use of AI to optimize battery performance at the laboratory level, shortening the cycle time for developing new battery materials will be of great value in advancement of battery technology and improving the life and specific capacity of batteries. At the level of Hybrid Electric Vehicle, for battery management systems, allowing real-time battery condition estimates using a data-driven approach offers a new perspective.

**Author information**

**Corresponding Author**

*E-mail: sunhui@cup.edu.cn.

**Notes**

The authors declare no competing financial interest.